\documentclass[11pt,a4paper]{article}
\usepackage[hyperref]{emnlp-ijcnlp-2019}
\usepackage{times}
\usepackage{latexsym}
\usepackage{algorithm}
\usepackage[noend]{algpseudocode}

\usepackage{url}

\usepackage{subfigure}
\usepackage{graphicx}
\aclfinalcopy % Uncomment this line for the final submission

\title{Visualizing Trends of Key Roles in News Articles}

\author{Chen Xia$^{1}$\thanks{\ \ Equal contribution.}, Haoxiang Zhang$^1$\footnotemark[1], Jacob Moghtader$^2$, Allen Wu$^2$, Kai-Wei Chang$^{1}$ \\
  $^1$University of California Los Angeles, 
  $^2$Taboola \\
  {\tt kasinxc@cs.ucla.edu}; 
  {\tt haoxiangzhx@gmail.com}; \\
  {\tt \{jacob.m, allen.wu\}@taboola.com}; {\tt kw@kwchang.net}
} 

\date{}

\begin{document}
\maketitle
\begin{abstract}

% There are tons of news generated every day reflecting the change of key roles, which include person, words and event names. Analyzing the trend of such key roles can efficiently obtain role relations and have a better understanding of the massive news data.

% Analyzing the trend of such key roles can efficiently obtain role relations and a better understanding of the massive news data.

There are tons of news articles generated every day reflecting the activities of key roles such as people, organizations and political parties. Analyzing these key roles allows us  to understand the trends in news. In this paper, we present a demonstration system that visualizes the trend of key roles in news articles  based on natural language processing techniques. Specifically, we apply a semantic role labeler and the dynamic word embedding technique to understand relationships between key roles in the news across different time periods and visualize the trends of key role and news topics change over time.

\end{abstract}

\section{Introduction}
Nowadays, numerous news articles describing different aspects of topics are flowing through the internet and media. Underneath the news flow, key roles including people and organizations interact with each other and involve in various events over time. With the overwhelmed information, extracting relations between key roles allows users to better understand what a key person is doing and how he/she is related to different news topics. To understand the action of key roles, we provide a semantic level analysis using semantic role labeling (SRL). To measure the trend of news topics, a word vector level analysis is supported using dynamic word embeddings.

% Semantic role labeling (SRL) extracts semantic roles from the news and  the biggest problem becomes how can people find a way to gather information efficiently from the unordered or even unstructured data? To better analyze the news trends, an intuitive way is to apply different kinds of natural language processing (NLP) techniques to extract key information from news. Semantic role labeling and word embeddings are two of these techniques. 

In our system, we show that a semantic role labeller, which identifies subject, object, and verb in a sentence, provides a snapshot of news articles. Analyzing the change of verbs with fixed subject over time can track the actions of key roles. Besides, the relationships between subjects and objects reflect how key roles are involved in different events. We implemented the semantic role analyzer based on the SRL model in AllenNLP, which formulates a BIO tagging problem \cite{he2017deep} and uses deep bidirectional LSTMs to label semantic roles \cite{allennlp}.

% For example, Donald Trump has been involved in many events and looking through all the news to summarize what he is doing requires a lot of human work. However, by applying SRL to the news data involving Donald Trump, we can efficiently extract frequent actions Donald Trump takes and whom he takes the actions to.

On the other hand, word embeddings map words to vectors such that the embedding space captures the semantic similarity between words. We apply dynamic word embeddings to analyze the temporal changes, and leverage these to study the trend of news related to a key role. For example, President Trump is involved in many news events; therefore, he is associated with various news topics. By analyzing the association between ``Trump'' and other entities in different periods, we can characterize news trends around him.  For example,  in February 2019, ``Trump'' participated in the North Korea-United States Summit in Hanoi, Vietnam. The word embedding trained on news articles around that time period identifies ``Trump'' is closely associated with ``Kim Jun Un'' (the President of North Korea) and ``Vietnam'' (the country hosted the summit).

% For example, the word ``Trump'' appears in many news articles. During or immediately after an important event, the word ``Trump'' may become more relevant to the terms related to this event. When the 2019 North Korea-United States Summit happened in Hanoi, Vietnam, we observed that ``Trump'' is closely connected to ``Kim Jung Un'' (the President of North Korea) and ``Vietnam'' (the country that hosted this summit).

% The two major news datasets are collected by Taboola. One is a topic specific Trump dataset containing 20,833 news titles involving \textit{Trump} from late April to early July 2018. The other is general 6 months data from October 2018 to March 2019. It contains approximately 3 million news titles and hierarchical topic clusters.

We create a system based on two datasets collected by Taboola, a web advertising company. 1) \emph{Trump dataset} contains 20,833 English news titles in late April to early July 2018. 2) \emph{Newsroom dataset} contains approximately 3 million English news articles published in October 2018 to March 2019. The former provides a controllable experiment environment to study news related to President Donald Trump, and the second provides a comprehensive corpus covering wide ranges of news in the U.S. Source code of the demo is available at \url{https://bit.ly/32f8k3t} and more details are in \cite{msthesisHaoxiang,msthesisChen}.
% and we conduct the following pre-processing. [then describe the text information we used].

% This paper introduces a system that applies semantic role labeling and dynamic word embeddings to visualize the news trend of key roles. News data from October 2018 to March 2019 is used to train and analysis. \textbf{ToDo: SRL} Word2vec is the embedding method and alignment is used to connect embedding models of different time periods. Also, n-gram phrases are embedded and absolute drift is used to capture the words that chagne the most.

% \textbf{@TODO have a paragraph describe the demo system. How it is implement what types of data is used to train and analysis.}

% \textbf{Our system is implemented in a python notebook.} 

\section{Related Work}

Various systems to visualize the transition of topics in news articles have been published. \citet{kawai2008using} detected news sentiment and visualized them based on date and granularity such as city, prefecture, and country. \citet{ishikawa2007t} developed a system called T-Scroll (Trend/Topic-Scroll) to visualize the transition of topics extracted from news articles. \citet{fitzpatrick2003breakingstory} provided an interactive system called BreakingStory to visualize change in online news. \citet{cui2010dynamic} introduced TextWheel to convey the dynamic natures of news streams. \citet{feldman1998trend} introduced Trend Graphs for visualizing the evolution of concept relationships in large document collections. Unlike these works, our analysis focuses on the key roles in news articles. We extract semantic roles and word vectors from news articles to understand the action and visualize the trend of these key roles.

% To understand the action of key roles, we provide a prospective of semantic level using semantic role labeling (SRL). To measure the trend of news topics, a word vector level analysis is supported using dynamic word embeddings.

% \section{Dataset} The two major news datasets are collected by Taboola. One is a topic specific Trump dataset containing 20,833 news titles involving \textit{Trump} from late April to early July 2018. The other is general 6 months data from October 2018 to March 2019. It contains approximately 3 million news titles and hierarchical topic clusters.

% We experiments on articles published in XX/XX/XX to XX/XX/XX from major U.S. news media collected by Taboola. The dataset consists of XXX news articles and we conduct the following pre-processing. [then describe the text information we used].

% \textbf{@TODO: 
% provide a few more details here like the size of the data set etc and what information (e.g., titles of the news etc). You can set the data is collected by Taboola. (We also need to ask Jun about how much information we want to provide here). 
% }

\section{System Overview}
To visualize the news trends, we apply semantic role analysis and word embedding techniques. 

\begin{figure}[t]
	\includegraphics[width=0.9\linewidth]{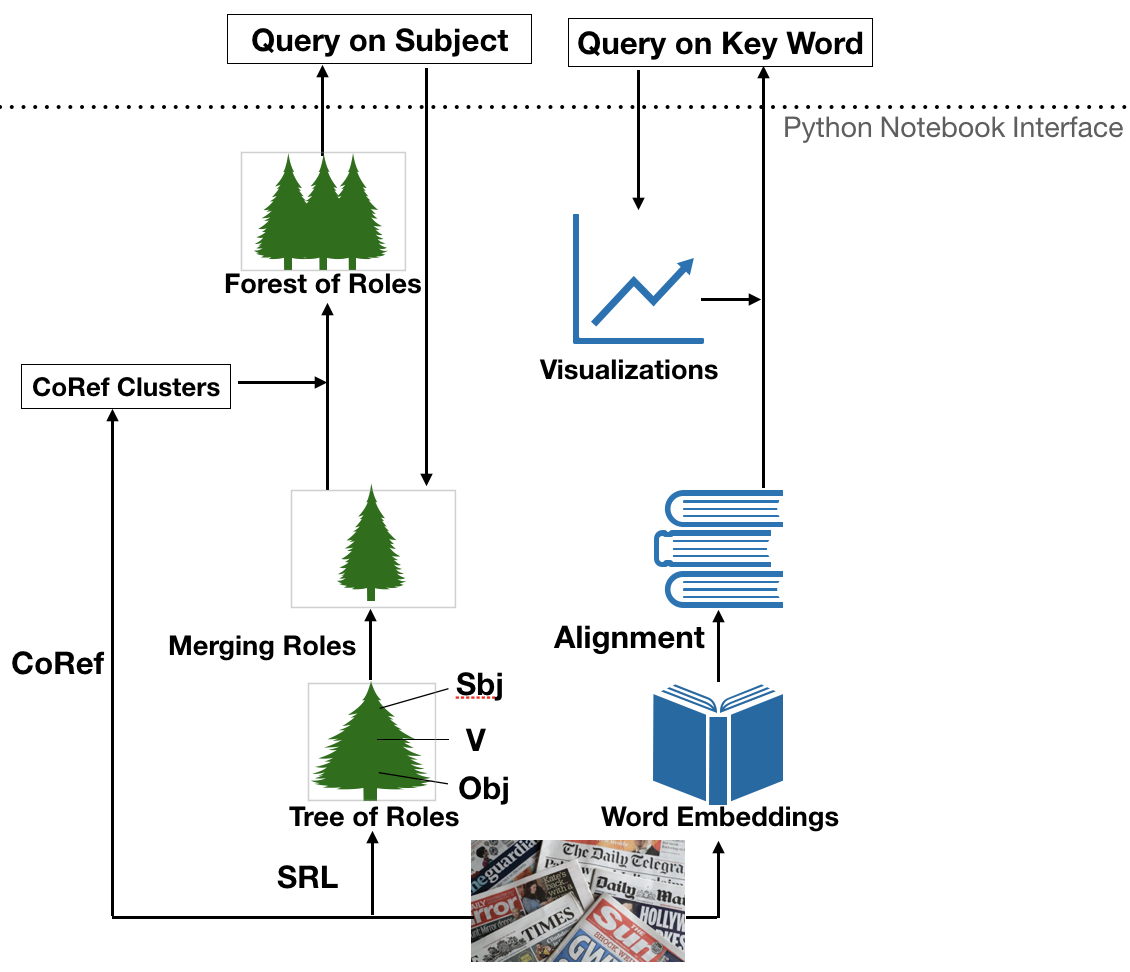}
	\caption{System Overview.}
	\label{fig:system overview}
\end{figure}

For semantic roles, we first construct a tree graph with subject as root, verbs as the first layer and objects as leaf nodes by extracting semantic roles with SRL \cite{allennlp}. Then we aggregate the tree graphs by collecting tree with the same subject and similar verb and object. Beyond applying simple string matching to identify same object and subject, we also apply a coreference resolution system (CoRef) to identify phrases refer to the same entity. As a result, we create a forest visualization where each tree represents the activities of a key role. 

% However, there probably has redundant information for similar verbs under same subject and similar objects under same verb. Then, we apply merge algorithms to simplify the tree. There are relevant trees whose subjects have the same referent. For each query on subject, we adopt coreference resolution (CoRef) to gather such relevant tree graphs to form a forest visualization. Tricks involving modifier, negative and lemmatizing verbs are also discussed later.

For word embeddings, we first train individual word vectors model for each month's data. However, there is no guarantee that coordinate axes of different models have similar latent semantics; therefore, we perform alignment algorithm to project all the word vectors into the same space. Once the embeddings are aligned, we are able to identify the shift of association between key roles and other news concepts based on their positions in the embedding space.

% In addition, we embed the n-gram phrases and define the absolute drift as the metric to detect words that change the most. For each query of a key word, we provide visualizations including 2D projection, table and line plot.

% \textbf{@TODO: You want to give an overview of the process before going to details. Say your demo system consists the following steps ....}

% \textbf{Describe the role of each step in the system (e.g., why you need to use coref why you need to merge entities ...)}
% \textbf{Then describe each step with example.} 

\subsection{Visualization by Semantic Roles}
% Our system firstly extracts semantic roles by a NLP library Allennlp \cite{allennlp} and then builds a tree structure with subject as root, verbs as the first layer and objects as leaf nodes. This only provides an exact match for subject of interest. For example, by searching \textit{Trump}, other names referring \textit{Trump} such as \textit{Donald Trump}, \textit{President Trump} are also wanted. Therefore, coreference resolution is applied to cluster tree graphs by root and generates a forest graph. 

% There are other merging challenges on verb-object and subject-verb edges in the graph. Tricks involving modifier, negative and lemmatizing verbs are also discussed as follows.

\begin{figure}[t]
    \centering
    \includegraphics[width=\linewidth]{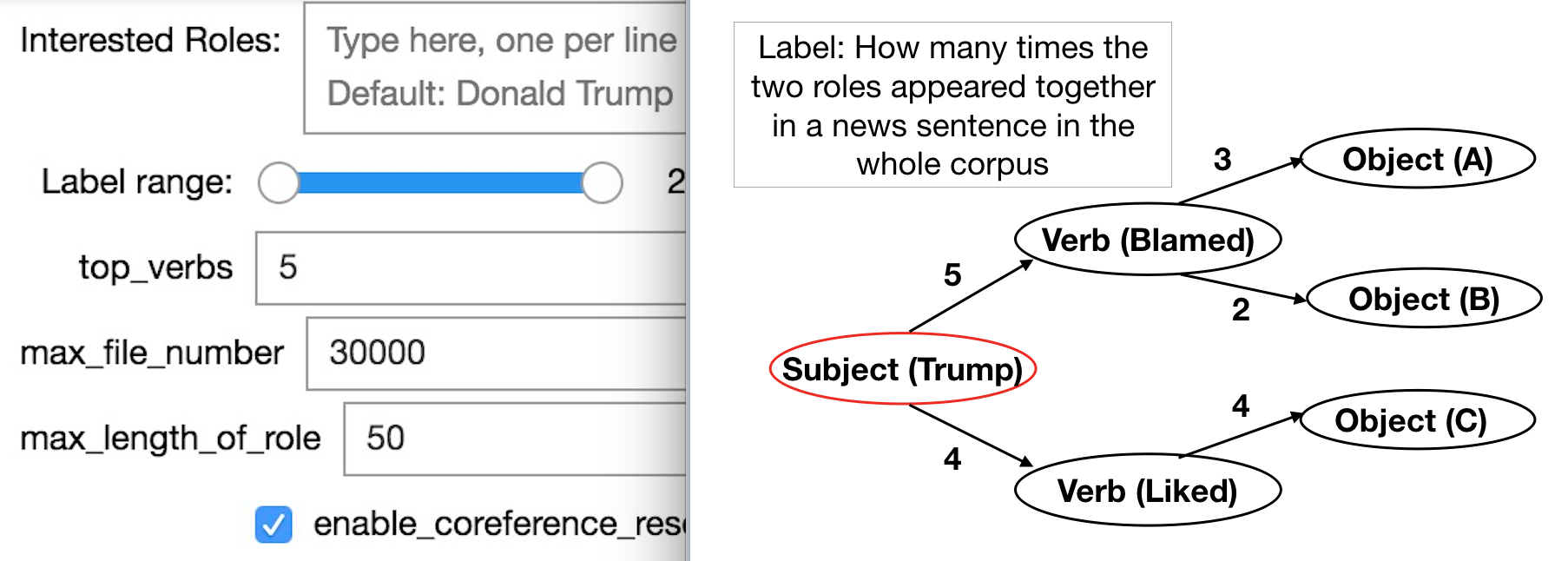}
	\caption{Tree Graph for Semantic Role Visualization.}
	\label{fig:design_srl_tree}
\end{figure}

\textbf{Tree Graph for Semantic Roles}
We provide users with a search bar to explore roles of interest. For example, when searching for \textit{Trump}, a tree graph is presented with \textit{Trump} as root. The second layer of the tree is all of the verbs labeled together with subject \textit{Trump}, e.g.,  \textit{blamed} and \textit{liked} in Figure~\ref{fig:design_srl_tree}. The edge label represents how many times two nodes, subject (e.g, \textit{Trump}) and Verb (e.g., \textit{liked}), appear together in a news sentence in the corpus. The edge label reflects the total number of semantic role combination in the given dataset, which depicts the importance of a news action.

% The leaf nodes are the objects of the predicate. From Figure~\ref{fig:design_srl_tree}, we can see that Trump has blaming \emph{person A} three times and another \emph{person B} twice. The subject-verb edge is sorted by labels so the tree graph actually depicts what the key subject often does.

% \begin{figure*}[t]
%     \centering
%     \includegraphics[width=0.6\textwidth,height=4cm]{fig/srl5.png}
% 	\caption{Tree Graph for Semantic Role Visualization.}
% 	\label{fig:design_srl_tree}
% \end{figure*}

% \begin{figure}[htb]
%     \centering
% 	\includegraphics[width=0.7\linewidth]{fig/interface.png}
% 	\caption{Tree Graph for Semantic Role Visualization.}
% 	\label{fig:design_srl_tree}
% \end{figure}

\textbf{Forest Graph for Semantic Roles}
In news articles, President Trump have different references, such as Donald Trump, the president of the United States, and pronoun ``he'' -- a well-known task, called coreference resolution. 
%Every time a subject of interest is typed in, a set of its other alias should be added as well. 
When generating semantic trees, the system should not look only for \textit{Trump} but also other references. To realize this, we preprocess the dataset with CoRef system \cite{lee2017end} in AllenNLP \cite{allennlp} and generate local coreference clusters for each news article. To obtain a global view, we merge the clusters across documents together until none of them shares a common role. A visualization demo for CoRef is also provided.

\begin{figure}[htb]
	\includegraphics[width=\linewidth]{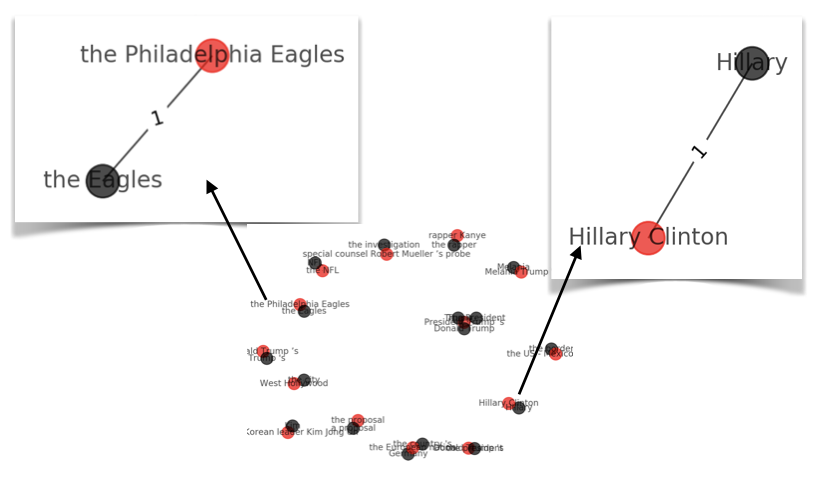}
	\caption{Coreference Resolution Clusters.}
	\label{fig:coref_tree}
\end{figure}
		
In Figure~\ref{fig:coref_tree}, the CoRef system clusters ``\textit{the Philladelphia Eagles}" with ``\textit{the Eagles}", and ``\textit{Hilary}" with ``\textit{Hilary Clinton}". The red nodes are center roles, which are representative phrases. For example, ``\textit{the Philladelphia Eagles}" and ``\textit{Hilary Clinton}" are the center roles of their corresponding cluster. 

We use the following three rules to determine which phrases are center roles. If phrases are tied, the one with longest length will be selected: \emph{LongestSpan} method selects the role with longest length. \emph{WordNet} method marks spans not in the WordNet \cite{miller1998wordnet} as specific roles. \emph{NameEntity} method marks roles in the name entity list generated by latent dirichlet allocation as specific ones. Both WordNet and NameEntity methods select the most frequent role as the center role.

\textbf{Merging Algorithms for Semantic Roles}
Finally, we use the following rule-based approach to merge trees with same referent subject by CoRef.
%Inside a tree, there also exist similar objects under a fixed verb, and similar verbs under a fixed subject. Such news graph contains redundancy. Our next challenge is how to merge objects and verbs reasonably. 

 \textbf{1) Merging Objects with the Same Verb} To better visualize the semantic roles, we merge objects with similar meaning if they are associated with same verb. To measure the similarity, we generate bag-of-word representations with \emph{TF-IDF} scores for each object. If the cosine similarity between the representations of two objects is larger than a threshold, we merge the two nodes.  
% apply the \emph{TF-IDF} and merge the nodes if the difference between objects   
% We merge objects with a certain verb into a new node and a new edge by \emph{TF-IDF}. Objects are merged together if the difference between their \emph{TF-IDF} score falls into a threshold set by user. The object with the highest \emph{TF-IDF} score is considered as a better representative of the topic and will be chosen as the new node. 
We then sum up the frequency weights on the edges of all merging objects to form a new edge.
 
\textbf{2) Merging Verbs with the Same Subject} Verbs like $believe$, $say$ and $think$ convey similar meanings. Merging such verbs can emphasize the key activities of the key roles. The similarity between verbs associated with the same subject is calculated by cosine similarity between word vectors using word2vec~\cite{mikolov2013distributed}. In particular, we merge two verbs if their cosine similarity is larger than a threshold.
% We also set up a cosine similarity threshold as a criterion to merge similar verbs. This merging process is conducted by union-find algorithm. 
By showing a certain range of edge labels, the system is also capable of filtering out verbs with extreme high or low frequency such as $say$, as these verbs carry less meaningful information.

% which benefits user to focus on representative actions.

% Term frequency $TF(t, d)$ and inverse document frequency $IDF(t, C)$ are calculated as follows:
% \begin{equation}
% TF(t, d) = \frac{O(t, d)}{\sum_{i \in d} O(i, d)}
% \end{equation}
% where $O(t, d)$ denotes the number of times term $t$ occurs in document $d$. 
% % Intuitively, the more representative the term $t$ is to the document, the higher $TF(t, d)$ score it can achieve.

% \begin{equation}
% IDF(t, C) = log(\frac{|C|}{|{d \in C : t \in d}| + 1})
% \end{equation}
% where $|C|$ denotes the number of documents in corpus $C$, and $|{d \in C : t \in d}|$ is the number of document in corpus that has term $t$ in it.
% \begin{equation}
% TF-IDF(t, d, C) = TF(t, d) \cdot IDF(t, C)
% \end{equation}

% Objects are merged together if the difference between their $TF-IDF$ score falls into a given threshold. The object with highest $TF-IDF$ score will be chosen to be the new node. 

% \textbf{\textit{Merging Verbs under 

% \subsubsection{}?

% Other semantic roles help convey the exact meanings of news. Modifier usually measures uncertainty. \textit{resign} and \textit{might resign} are totally different. We add them to verbs as an additional information for accuracy. 
\textbf{Modifier, Negative and Lemmatization} While our news analysis is mainly based on subject-verb-object relations, we also consider other semantic roles identified by the SRL model. For example, we include identification of modifier so that we can recognize the difference between ``resign'' and ``might resign''. We also add negation as an extra sentiment information. 
Verbs have different forms and tenses (e.g., win, won, winning). If we merge all verbs with the same root form, we can obtain a larger clusters and reduce duplicated trees. However, for some analysis, the tense of verbs are important. Therefore, we provide Lemmatizating as an option in our system. 

% Lemmatization focuses on verbs with different tense like \textit{winning} and \textit{won}, which share the stem. Merging them together removes redundant information, but lemmatized verbs lose information of tense. Whether lemmatizing a verb or not is an option and can be determined according to different user needs.

\subsection{Dynamic Word Embeddings}

%Static word embeddings assume that the meaning of any word does not change over time and train the entire corpus by one word embeddings model. For example, static word embeddings can definitely indicate that `bullish' and `bearish' are top neighboring words of the word `stock', but people cannot distinguish when the stock market is bullish or bearish. Dynamic word embeddings can help with this issue.

Dynamic word embeddings model align word embeddings trained on corpora collected in different time periods \cite{hamilton2016diachronic}. It divides data into time slices and obtains the word vector representations of each time slice separately. To capture how the trends in news change monthly, we train a word2vec word embedding model on news articles collected in each month. We then apply the orthogonal Procrustes to align the embeddings from different time periods by learning a transformation $\mathbf{R}^{(t)}\in {R}^{d\times d}$:
$$\mathbf{R}^{(t)}=\arg\min\nolimits_{\mathbf{Q}\top\mathbf{Q}=\mathbf{I}}\|\mathbf{W}^{(t)}\mathbf{Q}-\mathbf{W}^{(t+1)}\|,$$
where $W^{(t)}\in R^{d\times V}$ is the learned word embeddings of each month $t$ ($d$ is the dimension of word vector, and $V$ is the size of vocabulary).

% You may add the orthogonal procrustes equation and define the notations.  

\textbf{N-Gram} To represent named entities such as `white house' in the word embeddings, we treat phrases in news articles as single words. The max length of phrases is set as 4 to avoid large vocabulary size. 

% For example, \textit{chief investment officer of Bank of America Global Wealth and Investment Management} is a 12-word phrase found in one article. This phrase is verbose and is meaningless to encode into word embeddings because this phrase may only appear once among all time periods. The algorithm is simple and it will turn any phrase that is less than 5 words into one word, like treating `white house' as `white\_house'.
% \begin{algorithm}
% \caption{N gram}\label{euclid}
% \begin{algorithmic}[1]

% \State $\textbf{entitylen} \gets \text{length\ of\ }\textbf{entity}$
% \If {$\textbf{entitylen}<5$} 
% \State  $\textbf{joined} \gets \text{each\ word\ in\ }\textbf{entity}$ with ``\_"
% \State  $\text{replace\ the\ phrase\ in\ article\ by\ } \textbf{joined}$
% \EndIf

% \end{algorithmic}
% \end{algorithm}

\textbf{Absolute Drift} Inspired by \citet{rudolph2018dynamic}, we define a metric that is suitable to detect which words fluctuate the most relative to the key word $w_k$. Denote $cos(w_k, w_i, t)$ as the cosine similarities between the word $w_i$ and the key word $w_k$ at time $t$. For top n words close to $w_k$, calculate the absolute drift of each word $w_i$ by summing the cosine similarity differences.
$$drift(w_i)=\sum^T_{t=2}\!|\cos(w_k, w_i, t)\!-\!\cos(w_k, w_i, t-1)|$$ After finding meaningful words that fluctuate the most, cosine similarities between these words and $w_k$ of each month can be plotted to present possible useful interpretations.

\section{Case Studies}
\subsection{Semantic Roles}
\textbf{Action Tracking on Verbs}
We apply semantic role labelling model to \emph{newsroom dataset} collected from October 2018 to February 2019 on taxonomy: \textit{/sports/basketball} and search for subject \textit{LeBron\ James}, a basketball player. 

For each month, we generate the top frequent verbs from sentences where \textit{LeBron\ James} is marked as the subject. We found that the top verbs include ``Leave'', ``Score'' and ``Miss''. Example sentences include: "LeBron James \textbf{leave} the Cleveland Cavaliers", "LeBron James \textbf{score} points" and "LeBron James \textbf{miss} games".

We further show the ranking of these verbs in different months in Figure 4. As results show the verb ``leave'' ranks at the top around October due to an earlier announcement that Lebron James will leave the Cavaliers. However, the frequency falls in January.  
\begin{figure}[t]
	\centering
	\includegraphics[width=\linewidth]{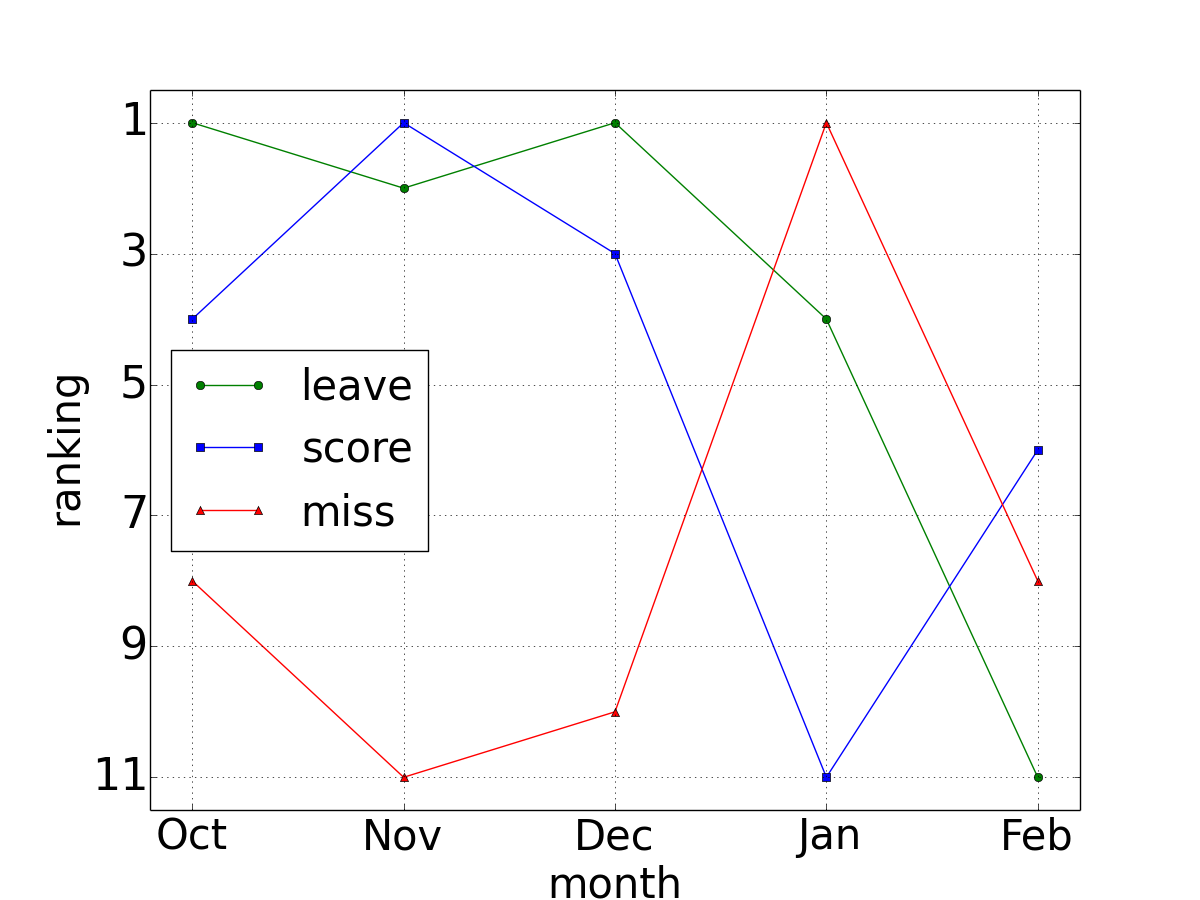}
	\caption{Action Tracking for \textit{LeBron James} }
	\label{fig:Action Tracking for LeBron James}
\end{figure}

 Meanwhile, news on \textit{LeBron James \textbf{miss} games} ranked first and the verb ``score" doesn't co-occur with LeBron James in January due to his injury.   

To explain the absence, we list the top $5$ frequent verbs are listed below. Verbs that occur with LeBron James only in December and January are colored in red.

\begin{table}[t]
	\centering
	\begin{tabular}{|p{0.7cm}|p{2.3cm}|p{3.4cm}|}
		\hline
		\textbf{rank} & \textbf{verbs for \textit{LeBron James}} & \textbf{fixed main objects}\\
		\hline
		1 & miss & games \\
		\hline
	\color{red}{2}&\color{red}{suffer}& \color{red}{a groin strain injury}\\
		\hline
		3  & make & no fixed main objects\\
		\hline
		4  & leave  & Cleveland Cavaliers\\
		\hline
		5  & lead & the team\\
		\hline
	\end{tabular}
	\caption{Verb Rankings for \textit{LeBron James} in January}
	\label {tbl:verb_rankings_for_lj}
\end{table}

From this analysis, we can see that \textit{LeBron James} was suffering the groin strain injury in January, causing his absence of the game.
 
\textbf{Breaking News Tracking on Objects}
We run our algorithm to analyze news article under the topic: \textit{/sports/basketball}, which has 75,827 peices of news title descriptions. We search \textit{Lakers} as subject in every month and sum up all the label weights on the edges between verb and object.
\begin{equation}
W(V, o | S = s) = \sum_{v \in V} W(v, o | S = s),
\end{equation}
where $W(v, o | S = s)$ denotes the weight on edges between all the verbs $v \in V$ and a specific object $o$ under certain subject $s$. 

We rank all objects based on Eq. (1) and the top 5 objects associated with the subject ``Lakers'' are: ``Davis'', ``James'', ``Game'', ``Ariza'', and ``Others''. We further show the pie chart to demonstrate the percentage of each object associated with ``Lakers'' in different months. 

\begin{figure}[t]
	\centering
    \includegraphics[width=\linewidth]{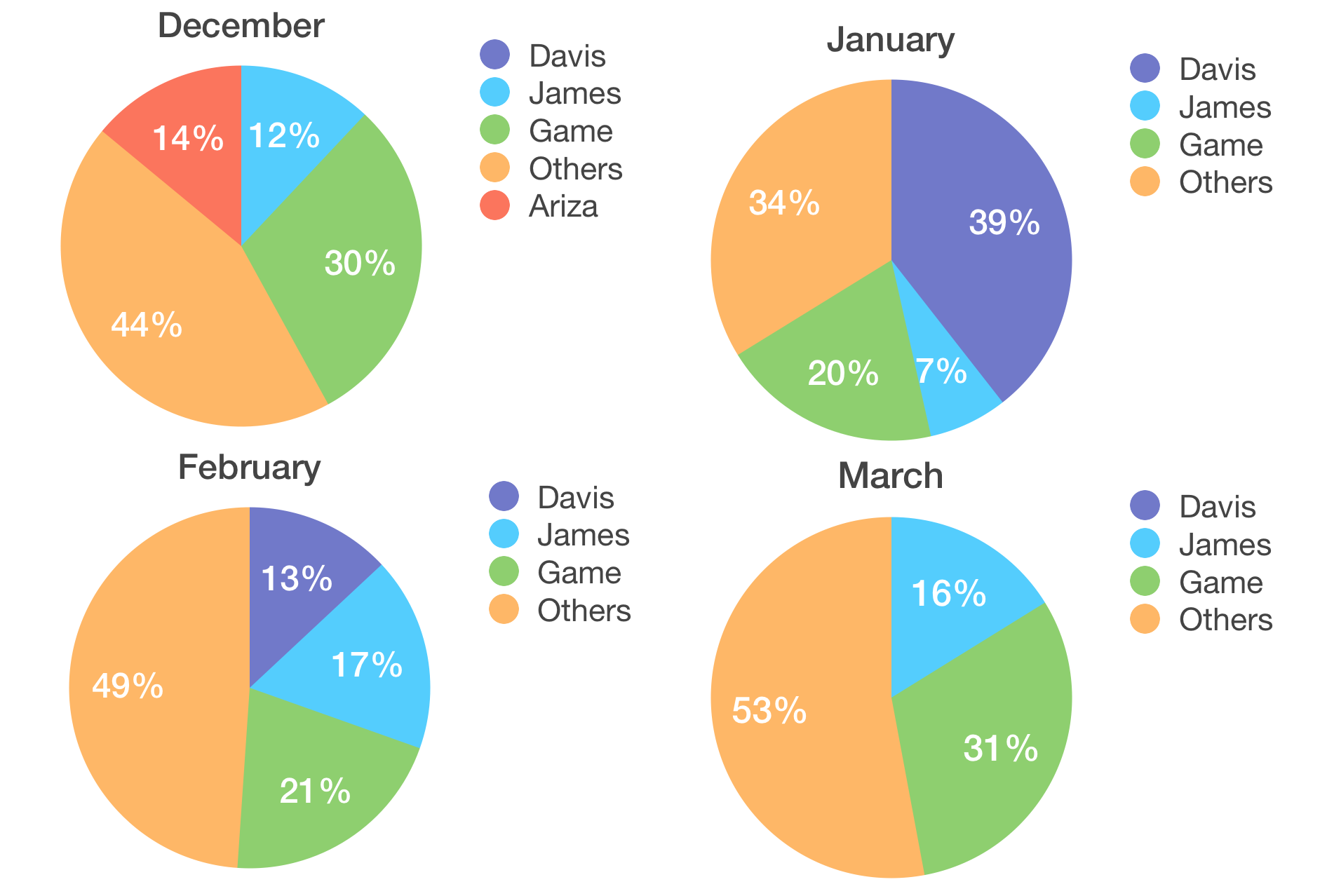}
	\caption{Breaking News Tracking on Trade Rumors. }
	\label{fig: Breaking News Tracking on Trade Rumors }
\end{figure}

The purple part in Figure~\ref{fig: Breaking News Tracking on Trade Rumors } shows that the number of news mentioning \textit{Anthony Davis} and \textit{Lakers} suddenly emerged and even beat \textit{James} and \textit{Lakers} in January but gradually decreased in February. The breaking news about Anthony and Lakers disappeared completely in March. The event happened in January and February was the trade rumors on $Davis$. After the trade deadlines, the topic eventually disappeared.

\subsection{Dynamic Word Embeddings}

\begin{table*}[t]
\centering
\begin{tabular}{|| c | c c c c||} 
 \hline
 Rank  & Dec 2018 & Jan 2019 & Feb 2019 & Mar 2019 \\ [0.5ex] 
 \hline\hline
 1 & los\_angeles\_lakers & los\_angeles\_lakers & los\_angeles\_lakers & los\_angeles\_lakers \\ 
 2 & lebron\_james & \textbf{pelicans} & lebron\_james & lebron\_james \\
 3 & lonzo\_ball & lebron\_james & clippers & clippers \\
 4 & clippers & lonzo\_ball & \textbf{pelicans} & kevin\_durant \\
 5 & brandon\_ingram & \textbf{anthony\_davis} & boston\_celtics & lonzo\_ball \\ 
 6 & kevin\_durant & cavs & kyle\_kuzma & lebron \\
 7 & \textbf{anthony\_davis} & boston\_celtics & tobias\_harris & giannis\_antetokounmpo \\
 8 & raptors & rockets & \textbf{anthony\_davis} & magic\_johnson \\
 [1ex] 
 \hline
\end{tabular}
\caption {Top 5 Words closest to the Word `lakers' in Each Month.}
\label {tbl:lakers}
\end{table*}

% \begin{table*}[t]
% \centering
% \begin{tabular}{|| c | c c c c c c||} 
%  \hline
%  Rank  & Oct 2018 & Nov 2018 & Dec 2018 & Jan 2019 & Feb 2019 & Mar 2019 \\ [0.5ex] 
%  \hline\hline
%  1 & la\_lakers	& lebron\_james & la\_lakers & la\_lakers & la\_lakers & la\_lakers \\ 
%  2 & storms & nor'easter & storms & \textbf{snow} & \textbf{snow} & \textbf{snow} \\
%  3 & tropical & \textbf{snowstorm} & rain & storms & storms & rain \\
%  4 & flooding & weather & flooding & rain & \textbf{snowfall} & \textbf{snowstorm} \\
%  5 & tornado & \textbf{snow} & \textbf{snow} & \textbf{snowfall} & weather & \textbf{blizzard} \\ [1ex] 
%  \hline
% \end{tabular}
% \caption {Top 5 Words closest to the Word `Storm' in Each Month (6-month data set)}
% \label {tbl:storm}
% \end{table*}

\textbf{2D Visualization} The t-SNE embedding method \cite{maaten2008visualizing} is used to visualize the word embeddings in two dimensions. First, given a word $w$ that we are interested in, the nearest neighbors of $w$ at different time periods are put together. Next, the t-SNE embeddings of these word vectors are calculated and visualized in a 2D plot.

\begin{figure}[htb]
\begin{center}
\includegraphics[width=\linewidth, angle=0]{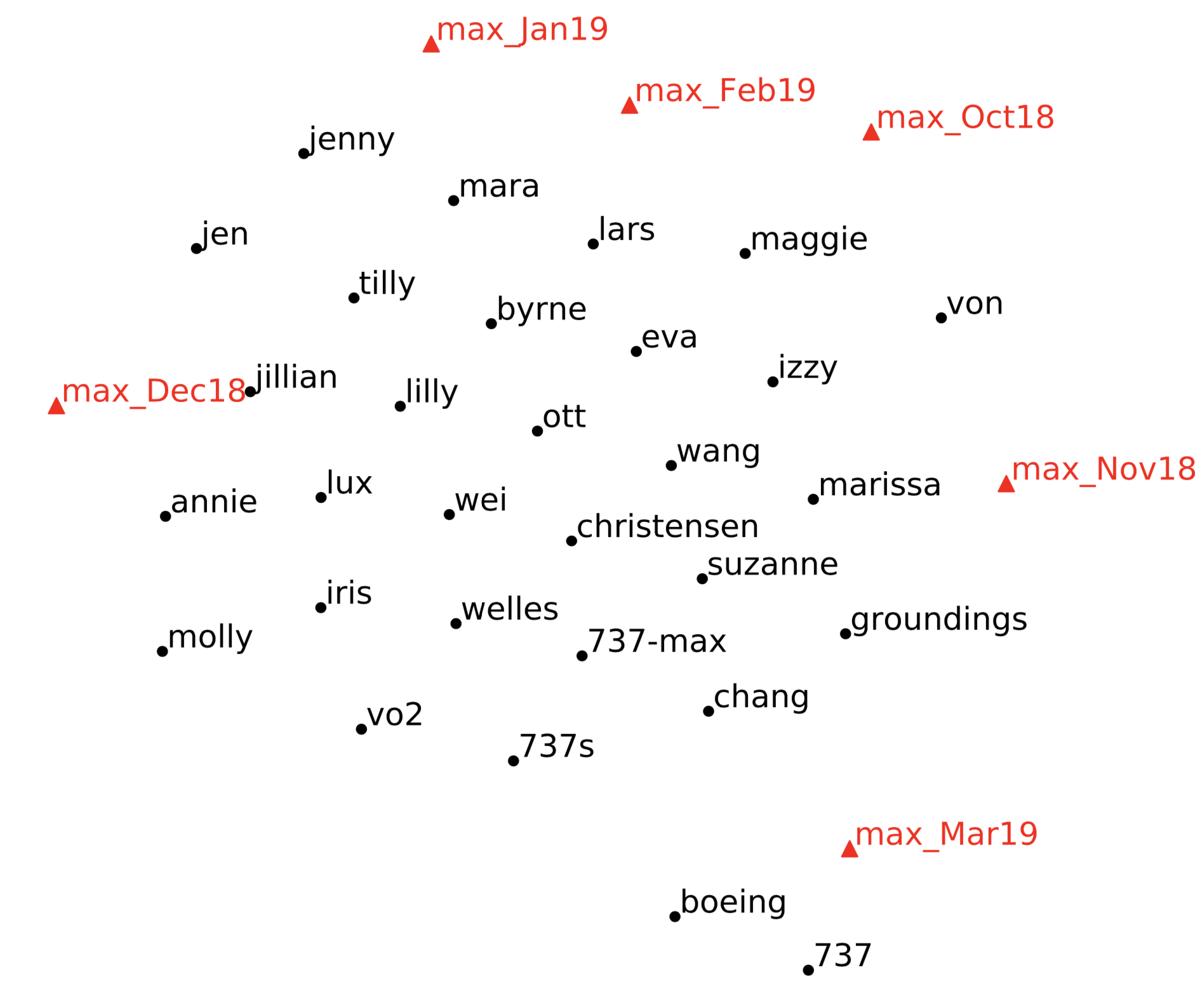}
\caption {Shifts of the Word `Max'.}
\label {fig:max}
\end{center}
\end{figure}

% Figure~\ref{fig:max} visualizes the shifts of the word 'max' from October 2018 to March 2019. The word `max1' is the embedding obtained from the October 2018 word embeddings model, `max2' from the November 2018 model, and so on.

On March 10 2019, the Boeing 737 MAX 8 aircraft crashed shortly after takeoff. After this fatal crash, aviation authorities around the world grounded the Boeing 737 MAX series. Figure~\ref{fig:max} shows that dynamic word embeddings capture this sudden trend change. In particular, before March 2019 (from when the `max\_Mar19' embedding is obtained), the word `max' was close to different people names. When the crash happened or afterwards, the word `max' immediately shifts to words such as `boeing', `737' and `grounding'. 

% or `xs' (this refers to the name iphone xs max)

\textbf{Top Nearest Nighbors} Listing the top nearest neighbors (words that have highest cosine similarities with the key word) of the key word $w$ inside a table also shows some interesting results. For example, Table~\ref{tbl:lakers} confirms with Figure~\ref{fig: Breaking News Tracking on Trade Rumors } that breaking news of \textit{Anthony Davis} and \textit{Lakers} happened because of the trade rumors.

\begin{figure}[htb]
\begin{center}
\includegraphics[width=\linewidth, angle=0]{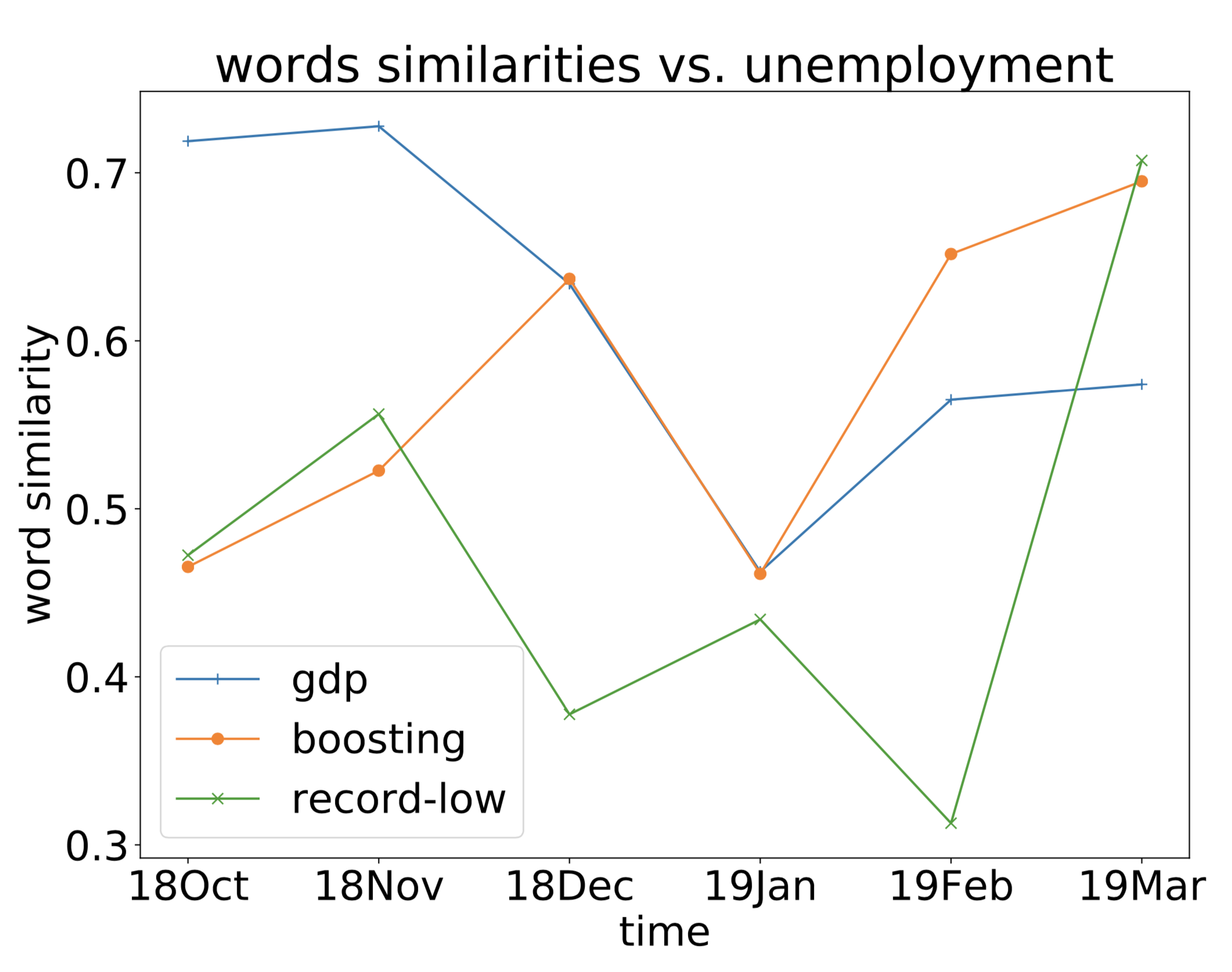}
\caption {Cosine Similarities with `Unemployment'.}
\label {fig:drift_unemployment}
\end{center}
\end{figure}

\textbf{Changing Words with Absolute Drift} Figure~\ref{fig:drift_unemployment} displays the cosine similarity changes with respect to `unemployment'. One thing we can infer from this figure is that as the economy (`gdp') shows a strong signal (`boosting') in the first quarter of 2019, the unemployment rate reaches a `record-low' position. According to National Public Radio, the first quarter's gross domestic product of U.S. grew at an annual rate of 3.2\%, which is a strong improvement compared to the 2.2\% at the end of last year. In addition, the Labor Department reported that 196,000 jobs were added in March, and the unemployment is near 50-year lows.

\section{Conclusion}
We presented a visualization system for analyzing news trends by applying semantic roles and word embeddings. We demonstrated that our system can track actions and breaking news. It can also detect meaningful words that change the most. Future work will focus on adding entity linking to subjects, providing information from other types of semantic roles. Also, we plan to work on qualitative assessment on the quality of the trends and other word embedding models like Glove \cite{pennington2014glove}.

\section{Acknowledgment}
This work was supported in part by a gift grant from Taboola. We acknowledge feedback from anonymous reviewers and fruitful discussions with the Taboola team at Los Angeles.

% \section{Acknowledgement}
% We appreciate Taboola company for providing the dataset. 
\bibliographystyle{acl_natbib}
\bibliography{emnlp-ijcnlp-2019}

\begin{thebibliography}{16}
\expandafter\ifx\csname natexlab\endcsname\relax\def\natexlab#1{#1}\fi

\bibitem[{Cui et~al.(2010)Cui, Zhou, Qu, Zhang, and Skiena}]{cui2010dynamic}
Weiwei Cui, Hong Zhou, Huamin Qu, Wenbin Zhang, and Steven Skiena. 2010.
\newblock A dynamic visual interface for news stream analysis.
\newblock In \emph{Proceedings of the first international workshop on
  Intelligent visual interfaces for text analysis}, pages 5--8. ACM.

\bibitem[{Feldman et~al.(1998)Feldman, Aumann, Zilberstein, and
  Ben-Yehuda}]{feldman1998trend}
Ronen Feldman, Yonatan Aumann, Amir Zilberstein, and Yaron Ben-Yehuda. 1998.
\newblock Trend graphs: Visualizing the evolution of concept relationships in
  large document collections.
\newblock In \emph{European Symposium on Principles of Data Mining and
  Knowledge Discovery}, pages 38--46. Springer.

\bibitem[{Fitzpatrick et~al.(2003)Fitzpatrick, Reffell, and
  Aydelott}]{fitzpatrick2003breakingstory}
Jean~Anne Fitzpatrick, James Reffell, and Moryma Aydelott. 2003.
\newblock Breakingstory: visualizing change in online news.
\newblock In \emph{CHI'03 Extended Abstracts on Human Factors in Computing
  Systems}, pages 900--901. ACM.

\bibitem[{Gardner et~al.(2018)Gardner, Grus, Neumann, Tafjord, Dasigi, Liu,
  Peters, Schmitz, and Zettlemoyer}]{allennlp}
Matt Gardner, Joel Grus, Mark Neumann, Oyvind Tafjord, Pradeep Dasigi, Nelson
  Liu, Matthew Peters, Michael Schmitz, and Luke Zettlemoyer. 2018.
\newblock Allennlp: A deep semantic natural language processing platform.
\newblock \emph{arXiv preprint arXiv:1803.07640}.

\bibitem[{Hamilton et~al.(2016)Hamilton, Leskovec, and
  Jurafsky}]{hamilton2016diachronic}
William~L. Hamilton, Jure Leskovec, and Dan Jurafsky. 2016.
\newblock \href {https://doi.org/10.18653/v1/P16-1141} {Diachronic word
  embeddings reveal statistical laws of semantic change}.
\newblock In \emph{Proceedings of the 54th Annual Meeting of the Association
  for Computational Linguistics (Volume 1: Long Papers)}, pages 1489--1501,
  Berlin, Germany. Association for Computational Linguistics.

\bibitem[{He et~al.(2017)He, Lee, Lewis, and Zettlemoyer}]{he2017deep}
Luheng He, Kenton Lee, Mike Lewis, and Luke Zettlemoyer. 2017.
\newblock Deep semantic role labeling: What works and what’s next.
\newblock In \emph{Proceedings of the 55th Annual Meeting of the Association
  for Computational Linguistics (Volume 1: Long Papers)}, pages 473--483.

\bibitem[{Ishikawa and Hasegawa(2007)}]{ishikawa2007t}
Yoshiharu Ishikawa and Mikine Hasegawa. 2007.
\newblock T-scroll: Visualizing trends in a time-series of documents for
  interactive user exploration.
\newblock In \emph{International Conference on Theory and Practice of Digital
  Libraries}, pages 235--246. Springer.

\bibitem[{Kawai et~al.(2008)Kawai, Fujita, Kumamoto, Jianwei, and
  Tanaka}]{kawai2008using}
Yukiko Kawai, Yusuke Fujita, Tadahiko Kumamoto, Jianwei Jianwei, and Katsumi
  Tanaka. 2008.
\newblock Using a sentiment map for visualizing credibility of news sites on
  the web.
\newblock In \emph{Proceedings of the 2nd ACM workshop on Information
  credibility on the web}, pages 53--58. ACM.

\bibitem[{Lee et~al.(2017)Lee, He, Lewis, and Zettlemoyer}]{lee2017end}
Kenton Lee, Luheng He, Mike Lewis, and Luke Zettlemoyer. 2017.
\newblock End-to-end neural coreference resolution.
\newblock \emph{arXiv preprint arXiv:1707.07045}.

\bibitem[{Maaten and Hinton(2008)}]{maaten2008visualizing}
Laurens van~der Maaten and Geoffrey Hinton. 2008.
\newblock Visualizing data using t-sne.
\newblock \emph{Journal of machine learning research}, 9(Nov):2579--2605.

\bibitem[{Mikolov et~al.(2013)Mikolov, Sutskever, Chen, Corrado, and
  Dean}]{mikolov2013distributed}
Tomas Mikolov, Ilya Sutskever, Kai Chen, Greg~S Corrado, and Jeff Dean. 2013.
\newblock Distributed representations of words and phrases and their
  compositionality.
\newblock In \emph{Advances in neural information processing systems}, pages
  3111--3119.

\bibitem[{Miller(1998)}]{miller1998wordnet}
George Miller. 1998.
\newblock \emph{WordNet: An electronic lexical database}.
\newblock MIT press.

\bibitem[{Pennington et~al.(2014)Pennington, Socher, and
  Manning}]{pennington2014glove}
Jeffrey Pennington, Richard Socher, and Christopher Manning. 2014.
\newblock Glove: Global vectors for word representation.
\newblock In \emph{Proceedings of the 2014 conference on empirical methods in
  natural language processing (EMNLP)}, pages 1532--1543.

\bibitem[{Rudolph and Blei(2018)}]{rudolph2018dynamic}
Maja Rudolph and David Blei. 2018.
\newblock Dynamic embeddings for language evolution.
\newblock In \emph{Proceedings of the 2018 World Wide Web Conference on World
  Wide Web}, pages 1003--1011.

\bibitem[{Xia(2019)}]{msthesisChen}
Chen Xia. 2019.
\newblock \href {https://escholarship.org/uc/item/0bv836gm} {Extracting global
  entities information from news}.
\newblock Master's thesis, University of California, Los Angeles, California,
  US, 6.

\bibitem[{Zhang(2019)}]{msthesisHaoxiang}
Haoxiang Zhang. 2019.
\newblock \href {https://escholarship.org/uc/item/9tp9g31f} {Dynamic word
  embedding for news analysis}.
\newblock Master's thesis, University of California, Los Angeles, California,
  US, 6.

\end{thebibliography}

\end{document}